# Clustering Without (Thinking About) Triangulation

**Denise L. Draper**
*ddraper@cs.washington.edu*
Department of Computer Science and Engineering
University of Washington, Seattle, WA 98195

## Abstract

The undirected technique for evaluating belief networks [Jensen *et al.*, 1990a, Lauritzen and Spiegelhalter, 1988] requires clustering the nodes in the network into a junction tree. In the traditional view, the junction tree is constructed from the cliques of the moralized and triangulated belief network: triangulation is taken to be the primitive concept, the goal towards which any clustering algorithm (*e.g.* node elimination) is directed. In this paper, we present an alternative conception of clustering, in which clusters and the junction tree property play the role of primitives: given a graph (not a tree) of clusters which obey (a modified version of) the junction tree property, we transform this graph until we have obtained a tree. There are several advantages to this approach: it is much clearer and easier to understand, which is important for humans who are constructing belief networks; it admits a wider range of heuristics which may enable more efficient or superior clustering algorithms; and it serves as the natural basis for an incremental clustering scheme, which we describe.

## 1 Introduction

Belief networks are directed acyclic graphs in which nodes represent uncertain variables and arcs between nodes represent probabilistic interactions between variables. A belief network is parameterized by providing, for each variable **X**, a conditional probability of that variable given its parents in the network, $P(\mathbf{X}|pa(\mathbf{X}))$. For variables with no parents in the network a simple prior probability $P(\mathbf{X})$ is provided. A belief network may be evaluated to give the marginal probability of all variables in the network, possibly conditioned on observations of values of some of the variables. There are several algorithms for evaluating belief networks, but it has been shown [Shachter *et al.*, 1994] that all known exact algorithms are equivalent to the undirected belief network evaluation technique of [Jensen *et al.*, 1990a, Lauritzen and Spiegelhalter, 1988]. This algorithm works in two stages: in the first stage a *junction tree* is constructed, and in the second stage, messages are propagated through the junction tree.

We will give a short review of triangulation and junction tree construction; for more detail see *e.g.* [Jensen, 1988], [Almond and Kong, 1991], [Kjærulff, 1990], [Jensen and Jensen, 1994].

Junction trees have traditionally been constructed by the following method:

1. "Moralize" the belief network by adding arcs connecting every pair of parents of any variable in the network, and dropping the directions on all arcs.
2. Triangulate the (now undirected) network by adding *fill arcs* until no chord-less cycles of length greater than three remain.
3. Create a tree whose vertices are the cliques of the triangulated graph and which are connected such that the *junction tree property* holds: if any two cliques contain a particular variable **V**, then every clique on the path between those two cliques must also contain that variable.

Triangulation is usually done by *node elimination*, which proceeds iteratively as follows:

1. Select a non-eliminated variable **X** to eliminate.
2. Add fill arcs connecting all the non-eliminated neighbors of **X** to one another (*i.e.* make the union of **X** and its neighbors fully connected).
3. Mark **X** as eliminated.

When all the variables have been eliminated, the graph is guaranteed to be triangulated.

The *cost* of a junction tree is the cost of doing the message propagation in that tree, and is proportional to the sum of the sizes of the "potentials" of each clique in the tree. The size of the potentials is the product of the number of states of each variable in the clique. Different choices of fill arcs can generate junction trees of radically differing cost; minimiz-ing the cost of a junction tree is NP-complete [Arnborg *et al.*, 1987]. The most effective known heuristic is the "minimum-weight" heuristic, which eliminates the variable whose non-eliminated neighbors form the clique with the smallest potential [Kjærulff, 1990].

One difficulty with node elimination, or with triangulation in general, is that it is very difficult to understand. The original graphical structure of the belief network doesn't help much: it is generally very difficult to determine visually if a graph is triangulated, and seeing the cliques of a triangulated graph is even more difficult, especially when they have large overlaps. Belief networks are often constructed by people, and given the complexity of belief network evaluation it may be assumed that people are interested in engineering their networks so as to reduce the cost of evaluation. The difficulty of visualizing the clustering process makes such engineering much more difficult.

The work of [Shachter *et al.*, 1994] and [Jensen and Jensen, 1994] underscores the fundamental importance that triangulation plays in all known belief network evaluation algorithms. But it does not follow that we must *think* in terms of triangulation—"hidden triangulation" may be just as effective as "overt triangulation."

In the remainder of this paper, we present a new framework for the construction of junction trees, which is not overtly based on triangulation. Section 2 introduces *cluster graphs*, which are a generalization of junction trees to multiply-connected graphs, and explains the general principles of transforming a cluster graph into a junction tree. Section 3 enumerates some of the transformations that can be used in this process, and Section 4 combines these transformations into two algorithms: the transformational equivalent of the node elimination algorithm and another completely new algorithm. Our original motivation in undertaking this research was to find an algorithm for incremental clustering which would allow the dynamic addition of variables and arcs to the belief network without forcing the recomputation of the entire junction tree, and Section 5 presents an algorithm for incremental clustering which arises quite naturally from our cluster graph framework. Section 6 presents some empirical results on our algorithms, and Section 7 concludes.

## 2 Cluster Graphs

A junction tree of a belief network $N$ is a graph $J$ whose vertices are *clusters* (sets of variables from $N$), and which has the following properties:

*Singly-Connected.* $J$ is a tree.

*Family Property.* For each variable $\mathtt{X}$ in the network $N$, there is some cluster $P$ in $J$ which contains the family of $\mathtt{X}$ (the union of $\mathtt{X}$ and its parents).

*Junction Tree Property.* For any two clusters $P$ and $Q$ in $J$ that contain a variable $\mathtt{X}$, every cluster on the path between $P$ and $Q$ must also contain $\mathtt{X}$.

Furthermore, with each edge between two clusters in $J$ there is associated a *separator*, which is defined to be the intersection of the two clusters.

We will generalize this definition by removing the first property, modifying the third, and changing the definition of separators.

A *cluster graph*[1] of a belief network $N$ is a graph $G$ whose vertices are clusters and which has the family property and the following *path property*, which is a modification of the junction tree property for multiply-connected graphs:

*Path Property.* For any two clusters $P$ and $Q$ both containing a variable $\mathtt{X}$, there exists *some* path between $P$ and $Q$, such that every cluster on that path contains $\mathtt{X}$, and the separator of every edge on that path contains $\mathtt{X}$.

With each edge between two clusters in $G$, there is associated a separator, which must be a *subset* of the intersection of the two clusters. We say that the edge *carries* the variables in its separator.

**Theorem 1:** A singly-connected cluster graph is a junction tree.

**Proof:** This is obvious except for the definition of separators. Consider two adjacent clusters $P$ and $Q$ in a singly-connected cluster graph $G$. The path property asserts that for every variable in $P \cap Q$, there exists some path between $P$ and $Q$ in which every edge carries that variable. But since $G$ is singly-connected, there is only one path between $P$ and $Q$, namely the edge that connects them. Thus this edge must carry $P \cap Q$. □

Finally, we note that a cluster graph for a belief network $N$ can be trivially constructed from $N$ by creating one cluster for each variable in $N$, containing the family of that variable, and connected in the same topology as $N$ (without the directions on the arcs). (If $N$ is singly-connected (a polytree), this cluster graph is also a junction tree.)

We can now describe our method for constructing junction trees: starting with the cluster graph for $N$ constructed as above, modify it using transformations which preserve the cluster graph properties, and which terminate when the cluster graph has been rendered singly-connected.

It should be clear that it is sufficient to resolve the multiply-connected components of the cluster graph separately: if each multiply-connected component of a cluster graph $G$ is transformed into a singly-connected subgraph (without adding more edges between components), then $G$ must also be singly-connected.[2] Furthermore, this restriction to multiply-connected components can be used recursively: if a transformation succeeds in rendering any set of clusters singly-connected in $G$, then it is not necessary to further consider transformations on those clusters.

[1] Note that our cluster graphs are different from the "junction graphs" of [Jensen *et al.*, 1990b]: junction graphs are used to find junction trees once the graph has been triangulated, and they have edges between every pair of overlapping cliques.

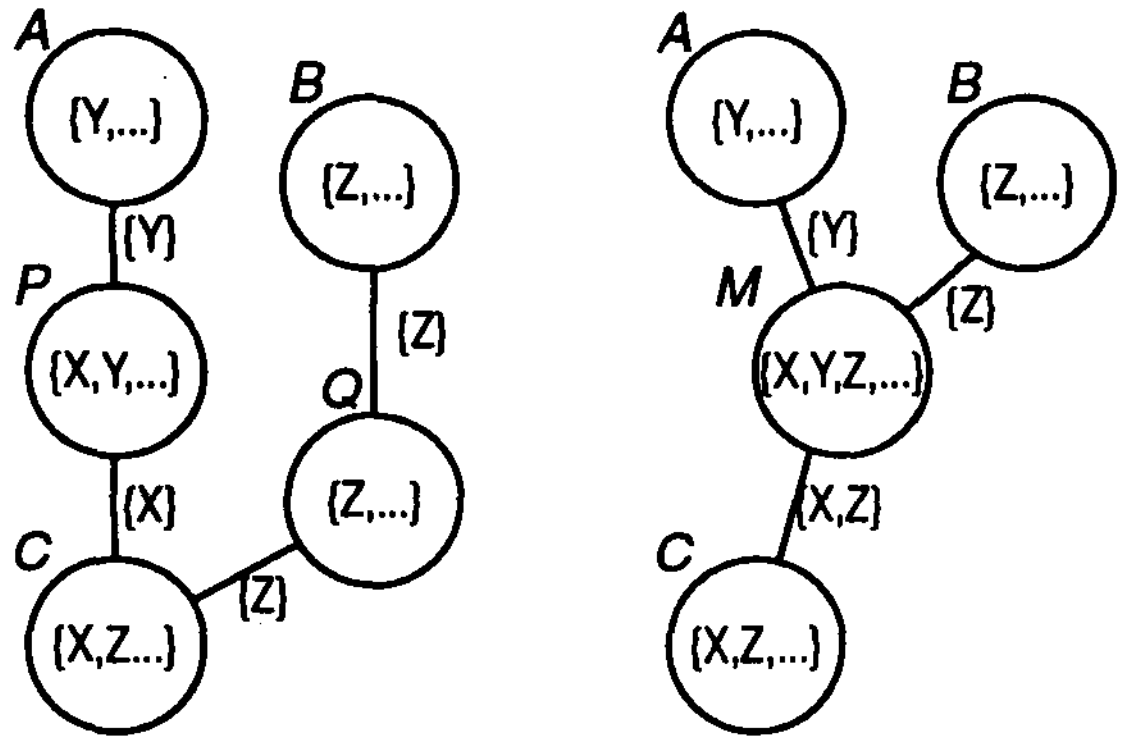

(a) (b)

Figure 1: The Merge transformation.

There are a wide variety of possible transformations and algorithms for using them; we will demonstrate some examples in the next two sections.

## 3 Transformations

A transformation is any operation that maps one cluster graph into another, preserving the family and path properties.[3] The transformations we have explored are very easy to grasp visually: they typically affect a small set of clusters by adding or deleting edges, adding variables to clusters, or merging clusters together.

**Merge.** Any two clusters can be merged by taking the union of their variables and the union of their edges to other clusters, as demonstrated in Figure 1. When two clusters $P$ and $Q$ are merged to create a new cluster $M$, and both had edges to some third cluster $C$, the edges $(P,C)$ and $(Q,C)$ must also be merged into a new edge $(M,C)$ by merging their separators. Pearl's clustering technique [Pearl, 1988] can be modeled as Merge transformations.

A special kind of merging is when one cluster is a superset of the other. From the triangulation perspective, this kind of merging happens automatically, since cliques are by definition maximal. In our scheme, such transformations must be done explicitly.

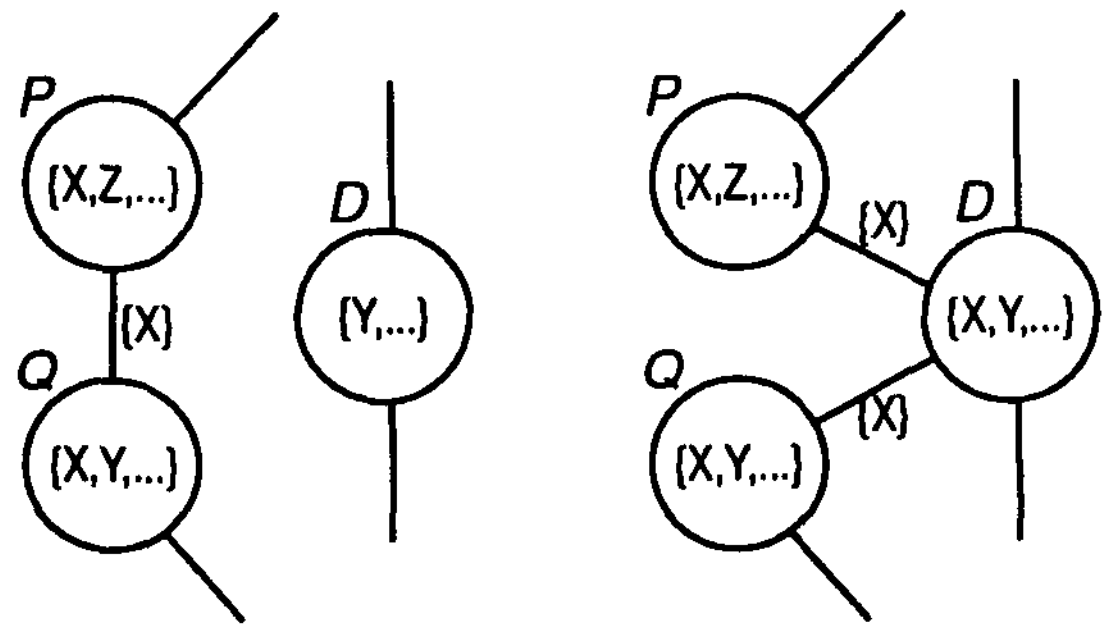

(a) (b)

Figure 2: The Steal-an-Edge transformation.

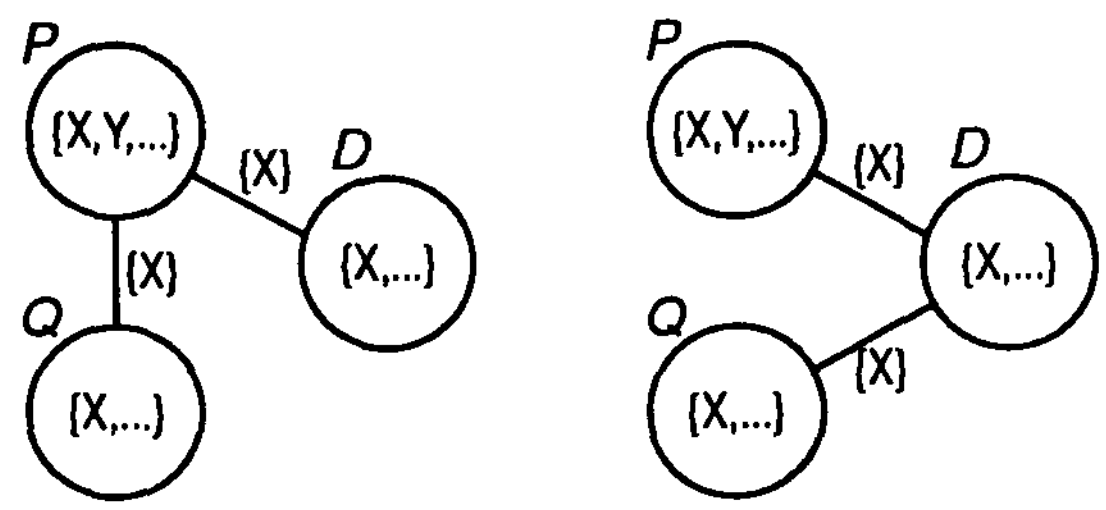

(a) (b)

Figure 3: The Slide transformation.

**Steal-an-Edge.** The Steal-an-Edge transformation is illustrated in Figure 2. An edge connecting two clusters $P$ and $Q$ is replaced by two edges which pass through a third cluster $D$. In order to retain the path property, variables carried by the edge $(P,Q)$ are added to $D$ if they are not already present. Note that the new edges $(P,D)$ and $(P,Q)$ need only carry the old separator of $(P,Q)$, even if there is a larger intersection between $D$ and the other clusters, as there is between $D$ and $Q$ in this example—if the path property held before this transformation, then there must be some other path carrying Y between $D$ and $Q$, and that path still exists after the transformation.

**Slide and Drop.** The Slide and Drop transformations can be seen as special cases of the Steal-an-Edge transformation where respectively one or both of the edges $(P,D)$ and $(Q,D)$ already exist. An example of Slide is shown in Figure 3; it is so-named because it resembles sliding one end of an edge from one cluster to a neighboring cluster. A Drop transformation occurs when there is a triangle, and one of its edges is simply deleted. In both cases, the "opposite" cluster ($D$) and the two edges to that cluster are augmented as necessary to carry the separator of the deleted edge.

**Collapse.** The Collapse transformation takes a simple cycle of clusters, deletes one edge from the cycle,

[2] In Section 4 we will give an even stronger result: it is possible, under certain reasonable conditions, to resolve $G$ by iteratively resolving *any* multiply-connected subgraph $S$ of $G$.

[3] We have not considered algorithms which might first violate then restore these properties, but this is clearly also possible.

and restores the path property by added the separator of the deleted edge to each other cluster and edge in the cycle. Following the argument of [Shachter *et al.*, 1994], this is the transformational equivalent of Loop-Cutset Conditioning.

**Node Elimination.** Eliminating a node can be modeled very elegantly as a transformation. The original definition of node elimination is: "select some variable (*i.e.* node) X, add fill-arcs as necessary to connect all its un-eliminated neighbors, and mark X as eliminated." Our Eliminate transformation takes as arguments the variable X and a set of clusters $S$ from which X is to be eliminated (here $S$ typically corresponds to the clusters containing unmarked variables). Since our clusters represent cliques (or subsets of cliques) in the underlying graph, the set of (unmarked) neighbors of X is the union of all variables in all the clusters in $S$ that contain X (the bold clusters in Figure 4(a)). Adding fill arcs to connect these neighbors is the same as merging those clusters in $S$ that contain X into a single cluster (*Elim-X* in Figure 4(b)). Marking the variable X as eliminated then translates to creating a second "buffer" cluster (*Buffer-X*) which contains all of the variables of the elimination cluster *except* X, and which also inherits all the edges from other clusters in $S$ that are incident to any of the merged clusters. By our construction, *Elim-X* is only connected to *Buffer-X* (within $S$), and cannot be part of any cycles remaining in $S$. Therefore *Elim-X* need not participate in any further transformations, and thus the variable X is effectively eliminated from consideration.

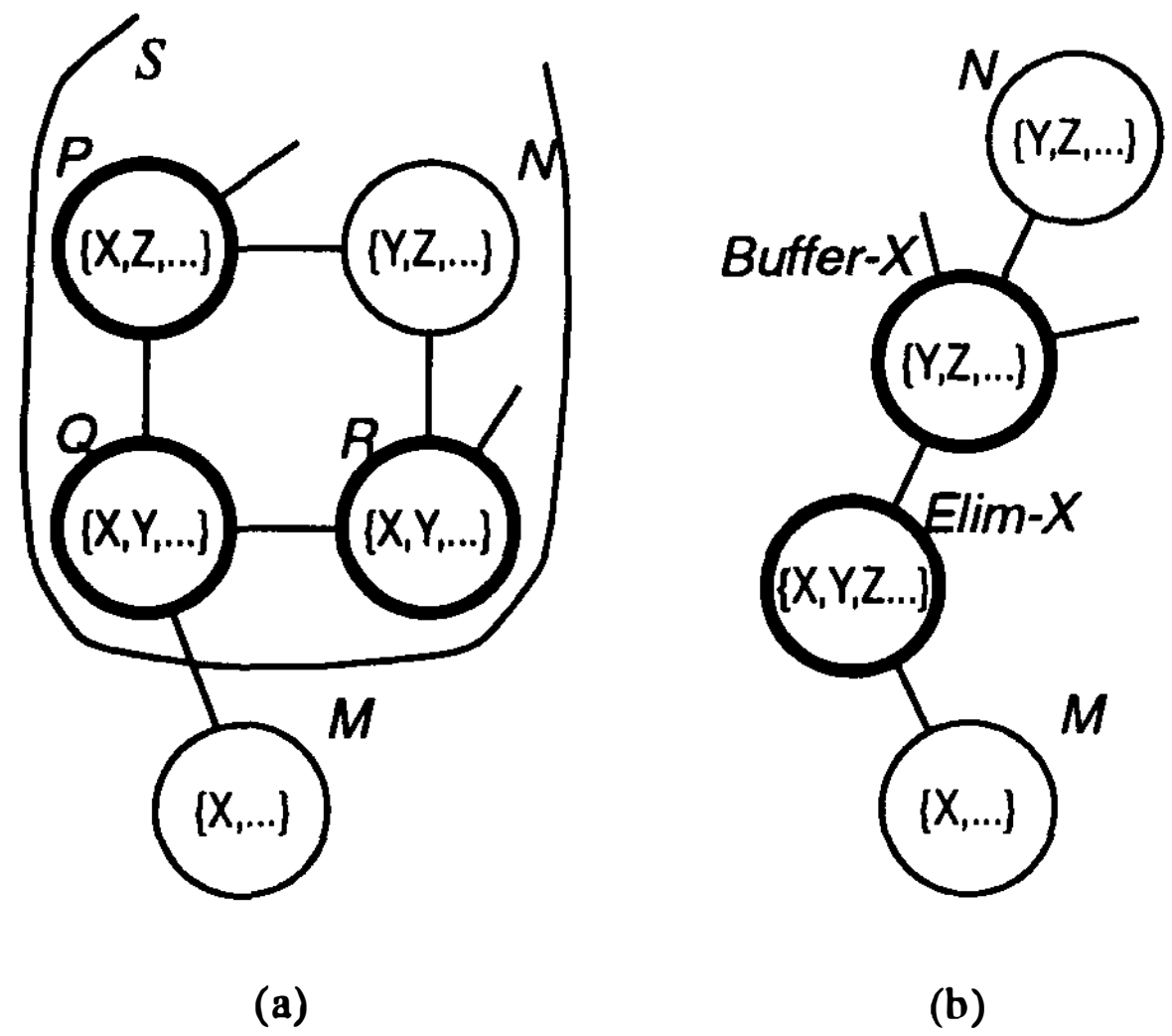

Figure 4: The Eliminate transformation eliminates the variable X from a set of clusters $S$ by merging the bold clusters in (a) into the two bold clusters (b).

When a variable X is eliminated from a set $S$ of clusters, it is only eliminated from those clusters in $S$; there may be other clusters outside of $S$ which still contain the variable X and are not merged into the elimination cluster (*e.g.* $M$ in Figure 4). Thus in order to retain the path property, any edges from outside $S$ to the merged clusters in $S$ that carry X must be connected to the elimination cluster rather than to the buffer cluster. (In fact, we simply migrate all edges between the merged clusters and clusters outside $S$ to the elimination cluster.)

In Section 2 we indicated that it is not necessary to consider the entire cluster graph when doing transformations: we can restrict our attention to a biconnected component of the cluster graph. When the set $S$ is a biconnected component, and a variable X is eliminated from $S$ but remains present outside of $S$ (as with X in $M$ above), we have reproduced the refinement to the elimination algorithm described by [Kjærulff, 1990, Theorem 3] (who further cites [Fujisawa and Orino, 1974] as its source). We wish to point out that this optimization arises quite naturally from the perspective of transformations and the need to preserve the path property.

**Dropping Spurious Variables.** Some transformations may result in a variable being present in a cluster where it serves no purpose: it is neither a member of any family in that cluster, nor is it being carried from one cluster to another. We call such a variable a *spurious* variable. (In order to detect spurious variables efficiently, we must note in each cluster which variables are part of families that must be preserved; obviously whenever clusters are merged, the new merged cluster must inherit the 'family' annotations from the old clusters.) Figure 3 shows an example of how this could arise with a Slide transformation; if the variable X is present in $P$ only so that $P$ can carry X between $Q$ and $D$, then after the transformation, X is spurious in $P$ (and on the edge $(P,D)$). Spurious variables can be generated by all the above transformations except Steal-an-Edge.

Spurious variables are dropped by removing them from the pertinent cluster and from the (at most one) incident edge that carries them. If either the edge or the cluster become empty as a result of this removal, they can be dropped from the cluster graph. We can either define the above transformations so they check for and remove spurious variables when they are created, or spurious variable removal can be defined as a separate transformation in its own right (in which case the transformation should operate recursively, possibly removing entire chains of spurious variables).

A more general notion of spuriousness is possible. Consider two clusters $P$ and $Q$ which need a variable X and are connected by two distinct paths, both carrying X. Unless every cluster on both of those paths also needs X, X could be removed from at least part of one of the paths without compromising the path property. A transformation which detected and removed X from some set of clusters and edges where it was not needed is possible, but the lack of locality of information might make it impractical.

For completeness, we will also describe the transformation corresponding to adding a single fill arc to the underlying moral graph of a belief network $N$. To add an arc from **X** to **Y**, a cluster $P$ that contains either **X** or **Y** is found. If cluster $P$ already contains both of **X** and **Y**, then there is nothing more to be done. If not, the remaining variable (say **Y**) is added to $P$ and the path property is restored by adding a new edge carrying **Y** from $P$ to any other cluster containing **Y**.

There are a wide variety of possible cluster graph transformations. It seems likely that a successful general clustering strategy would use only a few transformations, but there may also be transformations that are very useful in certain special cases.[4]

## 4 Algorithms

Now these transformations need to be put together into a sequence which will transform an arbitrary cluster graph into a junction tree. The issue of termination must be addressed: it should be clear that instances of the Slide transformation can undo each other, resulting in no progress. Cycles can also result from combinations of other transformations. We have not attempted a general categorization of terminating transformation sequences, but we will characterize one class of algorithms that function by repeatedly identifying and transforming multiply-connected subgraphs of the cluster graph.

We will show that we can construct an algorithm to transform a cluster graph $G$ into a junction tree starting with any algorithm that can transform a multiply-connected subgraph $S$ of $G$ into a singly-connected subgraph. The subgraph $S$ is not required to include all the edges in $G$ between clusters in $S$ (see Figure 5(a)).

In Section 3, transformations were described as deleting or adding edges to the graph. For the proof below, we will find it convenient to define *migrating an edge* to be deleting one edge and adding another with one of the same endpoints (*e.g.* deleting *(A,B)* and adding *(A,C)*). We also wish to generalize the idea of adding an edge so that we speak of "adding" an edge when the edge already exists in $G$, in which case the "new" edge is merged with the existing edge. For example, we will describe the Merge transformation as migrating edges, even when multiple edges become merged into one edge.

**Theorem 2:** An algorithm which functions by repeatedly finding some multiply-connected subgraph $S$ of a cluster graph $G$ (halting when there is no such $S$), and invoking a subroutine SUB on $S$, will terminate with $G$ transformed into a junction tree if the subroutine SUB obeys these properties:

1. SUB terminates.
2. SUB uses only transformations which preserve the family and path properties with respect to the entire graph $G$.
3. SUB transforms $S$ into a singly-connected subgraph.
4. The only effect SUB may have on clusters or edges outside of $S$ is that:
   (a) Edges not in $S$ but between clusters in $S$ may be deleted or migrated within $S$.
   (b) Edges between clusters in $G\backslash S$ and clusters in $S$ may be migrated, with only the endpoints in $S$ permitted to move.

**Proof:** $G$ is a junction tree if it is singly-connected and the family and path properties hold. That the family and path properties must always hold in $G$ follows from property (2) above. To show that $G$ must become singly-connected, first let us establish that it cannot become disconnected by an invocation of SUB on a subgraph $S$. Clearly $S$ cannot become disconnected, and any clusters in $G$ that were connected by paths outside of $S$ also remain connected. Let $R$ be the set of edges which connect clusters in $G\backslash S$ to clusters in $S$. If there are two clusters $P$ and $Q$ in $G\backslash S$ that were previously connected by some path passing through $S$, that path must have used some even number of edges from $R$. Suppose the first and last of those edges were *(G1,S1)* and *(S2,G2)*. After the subroutine, there must still be two edges *(G1,Sx)* and *(Sy,G2)* that are the possibly migrated versions of the old edges, and since $S$ is connected, there must be a path from $Sx$ to $Sy$, and thus between $P$ and $Q$, and thus $G$ remains connected. Now we will show that the quantity (# edges in $G$) − (# clusters in $G$) must decrease with each invocation of SUB; it must thus eventually reach $-1$, implying that $G$ is singly-connected. This is a simple algebraic proof once we have defined several quantities:

$$
\begin{aligned}
T &= G\backslash S = \text{the clusters in } G \text{ but not } S \\
& \quad \textit{before invoking } \text{SUB:} \\
n_T &= \text{the number of clusters in } T \\
n_S &= \text{the number of clusters in } S \\
e_T &= \text{the number of edges between clusters in } T \\
e_R &= \text{the number of edges between } T \text{ and } S \\
e_S &= \text{the number of edges in } S \\
k_S &= e_S - n_S \\
e_X &= \text{the number of edges between clusters in } S \\
& \quad \text{but not in } S \\
& \quad \textit{after invoking } \text{SUB:} \\
\Delta_S &= \text{the change in } n_S \\
\delta_R &= \text{the change in } e_R \\
\delta_X &= \text{the change in } e_X
\end{aligned}
$$

[4]An example is the "dynamic restructuring" transformation described in [Shachter *et al.*, 1994], which can be used to take advantage of the location of evidence in junction trees of a certain topology.

Initially, the total number of clusters in $G$ is $(n_T + n_S)$ and the number of edges is $(e_T + e_R + e_S + e_X)$. After SUB has been invoked, the number of clusters is $(n_T + n_S + \Delta_S)$ and the number of edges is $(e_T + (e_R + \delta_R) + (n_S + \Delta_S - 1) + (e_X + \delta_X))$. Putting all this together, we have:

$$\begin{aligned} & ((\#\text{ edges before}) - (\#\text{ clusters before})) - \\ & \quad ((\#\text{ edges after}) - (\#\text{ clusters after})) \\ = & \ ((e_T + e_R + (n_S + k_S) + e_X) - (n_T + n_S)) - \\ & \quad ((e_T + (e_R + \delta_R) + (n_S + \Delta_S - 1) + \\ & \qquad (e_X + \delta_X)) - (n_T + n_S + \Delta_S)) \\ = & \ 1 + k_S - \delta_R - \delta_X \end{aligned}$$

Since $k_S \geq 0$, and $\delta_R, \delta_X \leq 0$, the difference between the number of edges and the number of clusters in $G$ must decrease by at least one with each invocation of SUB, and thus $G$ must eventually become singly-connected. □

Now we describe two algorithms for transforming an arbitrary cluster graph $G$ into a junction tree. The first is the node elimination algorithm, and the second is a new algorithm.

### 4.1 Node Elimination

Given a set of clusters $S$, take the set of variables to be eliminated to be the union of the variables of each of the clusters in $S$. Choose some variable **X** to eliminate; when it has been eliminated, some subset of the clusters in $S$ are merged, and two new clusters, *Elim-X* and *Buffer-X*, are created. Recursively call Node-Elimination on $S \backslash \{Elim\text{-}X\}$. Note that it is not necessary to continue until *all* variables have been eliminated, only until $S$ is singly-connected.

If we let $S$ be $G$ in the original invocation of Node-Elimination, this is the traditional node elimination algorithm. Alternatively, by Theorem 2, we could create a new algorithm by iteratively invoking Node-Elimination on some other $S$.

### 4.2 Divide Loops

Divide-Loops iteratively finds a cycle in the set of nodes and recursively "transforms it away" by a subroutine we will call Divide-a-Loop. Divide-a-Loop uses Steal-an-Edge or Slide to recursively subdivide a cycle into smaller cycles (or into one smaller cycle and a "branch"), until eventually encountering cycles of length three, which are resolved by the Drop transformation. Figure 5 shows an example. Theorem 2 guarantees that Divide-Loops will correctly transform $G$ into a junction tree, since Divide-a-Loop clearly terminates, and also cannot affect edges or clusters outside the cycle $S$ on which it is invoked, except possibly to delete some other edges between clusters in $S$. Divide-Loops can be parameterized by the choice of which cycle to work on next, and Divide-a-Loop by which transformation to apply to which trio of clusters. We have tried several greedy heuristics, discussed in Section 6.

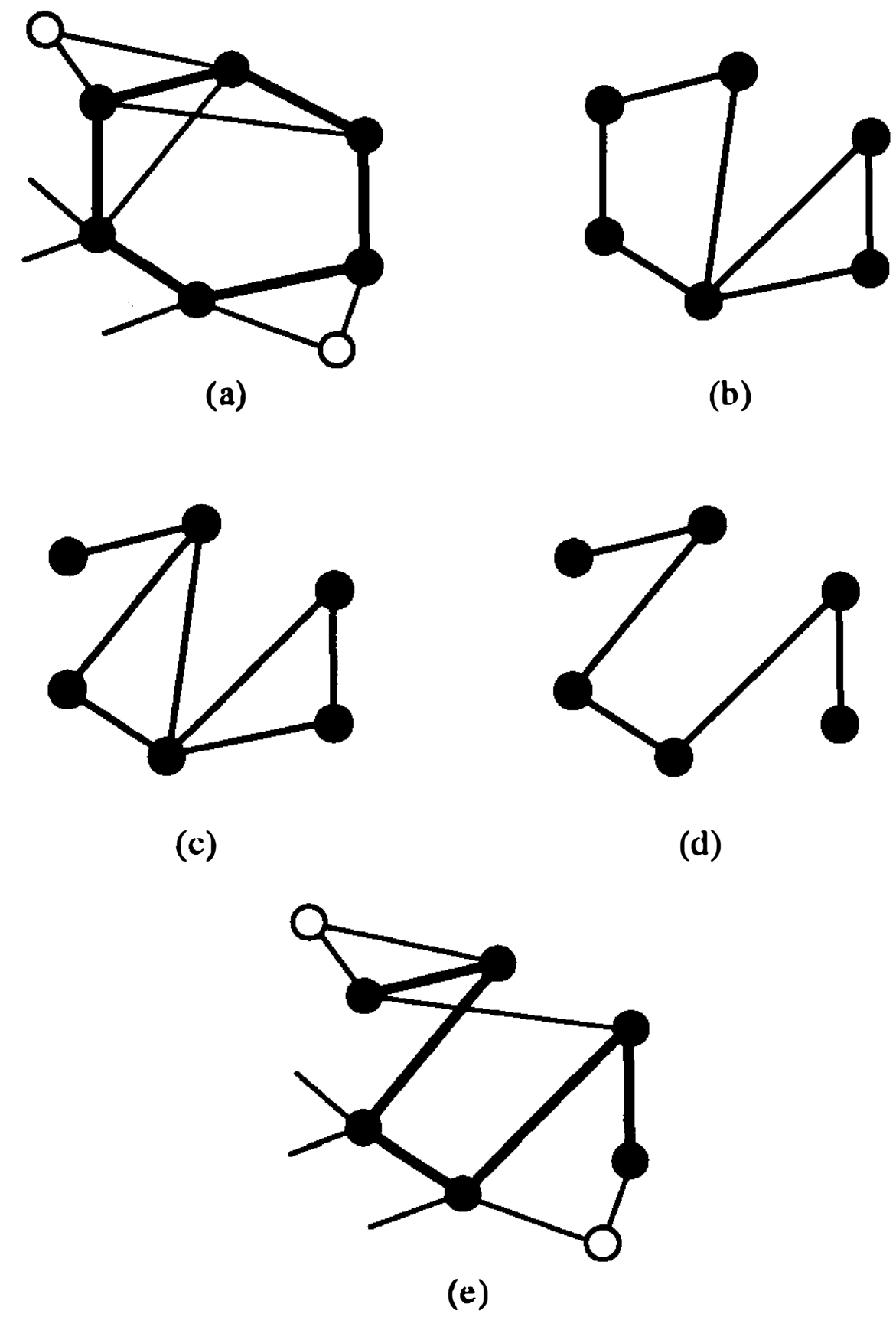

Figure 5: Divide-Loops transforms a graph into a tree by recursively subdividing each cycle.

### 4.3 Pre- and Postprocessing

In addition to procedures which render a subgraph $S$ singly-connected, we may also consider procedures on $S$ which transform it in other ways. *Preprocessing* is not aimed at making the subgraph singly-connected, but rather at simplifying it in ways we hope will improve the performance of the main algorithm (either by making it faster, or by causing it to generate a lower cost tree). *Postprocessing* takes a singly-connected subgraph and applies transformations to improve its cost. Pre- and postprocessing procedures can be applied globally to $G$ or individually to subgraphs $S$.

Free-Variable-Elimination is a preprocessing procedure which uses the Eliminate transformation to eliminate any variable that occurs in only one cluster in $S$. In node elimination, it is known that free variables can always be eliminated first without increasing the cost of the final junction tree [Rose *et al.*, 1976].

Merging redundant clusters (where one cluster is a subset of another) can be a preprocessing step. There is one subtlety: suppose that $P$ is a subset of $Q$, but some variables present in $P$ are only in $Q$ because $Q$ is carrying them between two other neighbors (*i.e.* they are not part of any family in $Q$). It might be disadvantageous to merge $P$ and $Q$ as a preprocessing step, because subsequent transformations might remove those variables from $Q$, if they are not merged. To avoid this, we require that the family variables of $P$ be a subset of the family variables of $Q$.

Merging redundant clusters is also valuable as a post-processing step. The definition of a junction tree given in Section 2 does not require that clusters must be cliques (that is, maximal), but clearly there is no advantage to retaining clusters which are not cliques.

In Section 3 (and Figure 3), we indicated that the Slide transformation might result in spurious variables, which could then be dropped. The combination of Slide and dropping spurious variables changes the cost of the cluster graph: the change is equal to the decrease in the size of the potential of the cluster which looses the edge ($P$ in Figure 3) minus the increase in the size of the cluster which acquires the edge ($D$). Note that when Slide is applied to a tree, the result is also a tree. Thus we have the postprocessing procedure Slide-Beneficially, which "juggles" the edges of the tree about, finding Slide transformations which reduce the cost of the tree by dropping spurious variables.

## 5 Incremental Clustering

If the structure of a belief network is modified dynamically, there are at least two reasons why it might be desirable to modify an existing junction tree incrementally rather than generating a new one from scratch: (1) If the network is very large, modifying an existing junction tree might be considerably less expensive than generating a new one. (2) Incremental modification should produce more stable results (*i.e.* more like the previous tree), which is important if the junction tree is being used by the network designer (especially if the designer is trying to optimize the junction tree itself).

Incremental clustering is very natural in our framework. We assume that the basic acts in modifying the belief network are the addition or deletion of an arc or a variable.

*To add a new variable.* Create a new cluster containing the variable (if the variable has parents, add arcs as follows).

*To add a new arc* X→Y. Find the cluster $P$ containing the family of the variable Y, and add X to $P$ and to the family. Find the cluster $Q$ containing the family of X, and if it is not the same as $P$, add an edge $(P,Q)$, or add X to the separator of the existing edge.

*To delete an arc* X→Y. Find the cluster $P$ containing the family of the variable Y, and remove X from the family of Y. If X has become spurious in $P$, it may simply be dropped from $P$. If X does not appear in any other family in $P$, but still is not spurious in $P$ (because $P$ carries X between more than one other adjacent clusters), we may *retract* X from $P$ by adding edges (carrying X) sufficient to connect all of $P$'s neighbors which contain X, and removing X from $P$ and all its incident edges. Retraction may be carried out recursively until the only edges carrying X are between clusters which have X as a member of some family.[5]

*To delete a variable.* Delete all the arcs between the variable and all other variables, then remove the family from its cluster (and the cluster itself, if it is now empty.)

If the modifications to the junction tree cause it to become multiply-connected, invoke some algorithm to transform it back into a tree. As with the transformations in Section 3, the correctness of these procedures follows from their maintenance of the family and path properties.

Incremental clustering might also be used when the network is modified in other ways. For example, a junction tree might be modified to optimize computation given the placement of evidence in the network. Or, if a particular query variable is identified, the junction tree could be modified to omit d-separated parts of the network.

## 6 Some Empirical Results

The transformations from Section 3 and the algorithms from Section 4 have been implemented in Common Lisp. We have run experiments on two networks, the Medianus-I and Medianus-II networks used for the study of triangulation techniques in [Kjærulff, 1990], and on four randomly generated networks. Medianus-I has 43 variables and 66 arcs, Medianus-II has 58 variables and 79 arcs, and the names of the randomly generated networks (see Figure 6) indicate their number of variables and arcs respectively. All the networks have non-binary variables. We performed two kinds of tests: "whole" tests, in which an algorithm was applied to an entire network (actually, to the biconnected components of the original cluster graph), and incremental tests where the network was built up (contiguously) by adding arcs one by one, and calling one of the transformation algorithms whenever the cluster graph became multiply-connected. (We have not yet experimented with retraction.) For each kind of test, we ran the algorithms several times, randomly breaking any ties (*e.g.* generated by the heuristic ranking of which transformation to perform next).

For Node-Elimination we used the min-weight heuris-tic, and Merge-Redundant-Clusters as a postprocessing step (we also tried Slide-Beneficially, but it never helped). Quite a number of parameterizations of Divide-Loops are possible: choices of which pre- or postprocessing steps to apply (and whether to apply them to each cycle individually, *i.e.* with each call to Divide-a-Loop, or to apply them to the entire graph), which cycle to work on next (we picked a cluster at random, and tried finding the cycle according to shortest length, lowest cost, or a weighted combination of shortest length and lowest cost), and which transformation to use to divide a cycle (minimize increase in any cluster cost, minimize increase in overall cost, minimize increase in cluster degree, and combinations thereof). The results in Figure 6 show three different parameterizations of Divide-Loops (indicated by D, D2 and ID), which performed best, respectively, on "whole" experiments on the real networks, the "whole" experiments on the artificial networks, and the incremental experiments on all networks.

[5]To completely reverse the operations which generated a junction tree, it is also necessary to be able to "take apart" clusters into their component families (connected by sufficient edges to guarantee the path property, naturally). Whether or when to do this would be a matter of heuristics.

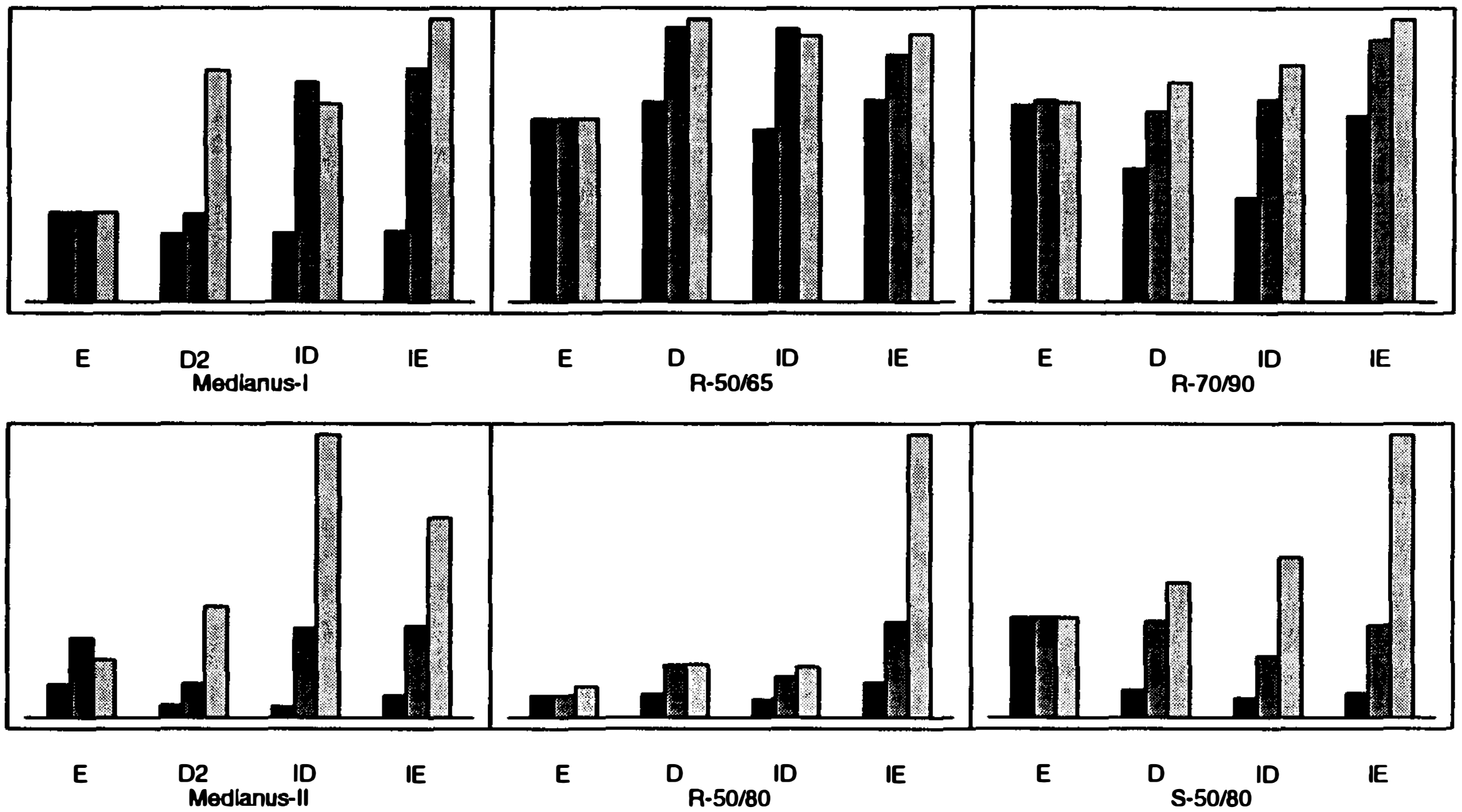

Figure 6: Comparison of Node-Elimination (E), Divide-Loop (D and D2), Divide-Loop applied incrementally (ID), and Node-Elimination applied incrementally (IE) on six networks. For each algorithm, the bars represent, in order, the minimum, median and average cost of the junction tree (sum the sizes of the cluster potentials) over 20 (or more) runs.

**D.** Preprocessing: free variable elimination on each call to Divide-a-Loop. Choice of cycle: weighted combination of shortest and cheapest (prefers shortest unless its very expensive). Choice of next transformation: minimize the increase in cluster cost. Postprocess the entire graph when finished by Slide-Beneficially followed by Merge-Redundant-Clusters.

**D2.** The same as **D**, except that choice of cycle is shortest cycle.

**ID.** The same as **D**, except that the postprocessing is done on each call to Divide-a-Loop, and does not include merging.

Probably the most surprising result was that merging redundant clusters is a very bad idea, except after all other processing has been done (and since there is no "last" in incremental clustering, it is best not to do it at all; but the results seem to show that this does not significantly affect the cost of the result).

The strongest impression from our "whole" experiments is that Divide-Loops has high variance, no matter what choice of parameters is used. In some cases the median may be lower than node elimination, but the average is always higher (but also, the minimum can be much lower). Since Divide-Loops works on only a single cycle at a time, it is perhaps not surprising that its performance is substantially the same when applied incrementally as when applied "whole" (except for Medianus-II). Node-Elimination, on the other hand, performs generally worse when applied incrementally, especially on the denser networks. It also appears that Divide-Loops competes more favorably with Node-Elimination on the denser networks than on the sparser networks.

Overall, these results indicate that while we have not found an algorithm which is superior to Node-Elimination, we have found a new algorithm, based on different principles, which is comparable. This encourages us to believe that our clustering framework may yield other good algorithms in the future.

## 7 Conclusion

Not long after beginning this research, we implemented a simple graphical interface to demonstrate the effect of some transformations. With this interface, we quickly discovered the existence of spurious variables, and were able to rule out some heuristics as ineffective. The graphical interface, and the simple visual nature of the transformations, greatly aided our understanding. We feel that this simplicity should be equally beneficial to belief network designers, making it easier to understand the computational properties of networks.

Further, viewing clustering as a process of transformations opens up a wide vista of possible new heuristic approaches to clustering, some of which may prove superior to known methods. However there are so many possible transformations and algorithms to employ them that we have something of an embarrassment of riches—there are simply too many possibilities to explore. Analysis revealing some organizational properties of this space would be quite helpful—for example, what sets of transformations are complete in the sense that they can generate all minimal triangulations?

In summary, we have presented a new framework for clustering algorithms, based on transformations of a cluster graph rather than on triangulation. We have demonstrated a set of transformations within this framework, and presented a new algorithm based on those transformations. We have also demonstrated an algorithm for incrementally clustering a dynamically changing network.

## Acknowledgments

We thank Uffe Kjærulff for providing us with the Medianus-I and Medianus-II networks, and Mike Williamson for insightful comments on the paper. This research was funded by National Science Foundation Grant IRI-9008670.